\pdfoutput=1

\documentclass[11pt,twocolumn,letterpaper]{article}

\usepackage[]{ACL2023}

\usepackage{times}
\usepackage{latexsym}
\usepackage{algorithm}
\usepackage{algpseudocode}

\usepackage[T1]{fontenc}

\usepackage[utf8]{inputenc}

\usepackage{microtype}

\usepackage{inconsolata}
\author{Shresth Grover \\
  IIT Kanpur \\
  \texttt{shrgo@iitk.ac.in} \\\And
 Vibhav Vineet \\
 Microsoft Research \\
  \texttt{vibhav.vineet@microsoft.com} \\\And
  Yogesh S Rawat\\
  CRCV, University of Central Florida\\
  \texttt{yogesh@crcv.ucf.edu}}
\title{
Navigating Hallucinations for Reasoning of Unintentional Activities
}

\begin{document}

\maketitle

\begin{abstract}

In this work we present a novel task of understanding unintentional human activities in videos. We formalize this problem as a reasoning task under zero-shot scenario, where given a video of an unintentional activity we want to know why it transitioned from intentional to unintentional. We first evaluate the effectiveness of current state-of-the-art Large Multimodal Models on this reasoning task and observe that they suffer from hallucination. We further propose a novel prompting technique, termed as Dream of Thoughts (DoT), which allows the model to navigate through hallucinated thoughts to achieve better reasoning. To evaluate the performance on this task, we also introduce three different specialized metrics designed to quantify the models reasoning capability.
We perform our experiments on two different datasets, OOPs and UCF-Crimes, and our findings show that DOT prompting technique is able to outperform standard prompting, while minimizing hallucinations. 
\end{abstract}
\section{Introduction}
\label{sec:intro}
 
Automatic understanding of human activities in videos is a challenging problem with a lot of real-world applications in domains such as healthcare, security, robotics, and elderly assistance. In past few years, we have seen an impressive progress in recognizing  intentional human activities in videos \cite{kong2022human}. However, human beings are prone to making mistakes and activities can be unintentional in real-world scenarios. Recognizing unintentional activities is  important \cite{Epstein_2020_CVPR}, but it is also important to understand the reasoning behind their occurrence. This can be useful for correcting mistakes and any damage control. Motivated by this, in this work we focus on finding the reasoning behind unintentional activities in videos. 
Recently developed multimodal foundation models have shown impressive capabilities across a range of tasks with strong generalization capabilities for zero-shot scenarios \cite{zhu2023minigpt, alayrac2022flamingo, li2023blip, liu2023visual, damonlpsg2023videollama, maaz2023video, li2023videochat}. We first study the reasoning abilities of existing Large Multimodal Models (LMMs) using prompting to determine the intentionality of actions as we transition to unintentional states. Our analysis reveal that conventional prompting techniques suffer from hallucinations and does not perform well in reasoning about the transition into unintentional activities. We also noticed that even when model is able to identify that the transition from intentional to unintentional has occurred it frequently provided very generic reasons without using the visual context to the fullest extent. Although chain of thoughts \cite{wei2022chain} prompting provides a framework to obtain specific reasons not just generic ones, it also suffers from hallucinations when trying to reason over unintentional activities. 
To mitigate the effect of hallucinations and improve the reasoning over unintentional activities, we propose a multi-step solution. Our solution relies on two key observations; 1) if we let a model hallucinate multiple times, some of the responses might be correct, and 2) multiple-choice questions helps guide the model to find the right answer. We build upon these observations and propose a novel approach termed Dream of Thought (DoT) style prompting. We use the models hallucinations and present to the model as multiple choices and let the model navigate through these choices and provide correct reasoning.

We experiment with two different datasets, OOPs \cite{Epstein_2020_CVPR} and UCF-Crimes \cite{sultani2018real}, where OOPs focus on unintentional activities in daily life and UCF-Crimes focus on anomalous activities. With extensive evaluations we demonstrate the effectiveness of DoT prompting over simple prompting and chain of thoughts prompting. We make the following contributions in this work,
\begin{itemize}
\itemsep0em 
    \item We present a novel problem that focuses on reasoning about the transition of an activity from intentional to unintentional.
    \item We study the capability of existing LMMs and prompting techniques for this task and also provide a novel Dream of Thoughts (DoT) reasoning-based mechanism which outperforms existing methods. 
    \item We provide three different evaluation protocols, $rm_{MCQ}, rm_{LLM}$, and $rm_{FIB}$, for response matching (rm)  
    which quantifies the reasoning capability of models for this task.
    
\end{itemize}

\section{Related works}
\textbf{Large generative models} 
The field of large language models (LLMs) has significantly evolved in recent years, with advanced models like GPT \cite{brown2020language}, LLaMA \cite{touvron2023llama}, ChatGPT \cite{OpenAI2023ChatGPT}, and BARD \cite{Google2023BARD}. These models excel at generalizing across various tasks.Emerging Large Multimodal models, derived from these foundational LLMs, are now being explored for vision tasks. Examples include MiniGPT \cite{zhu2023minigpt}, Open Flamingo \cite{alayrac2022flamingo}, BLiPv2 \cite{li2023blip}, and LLaVA \cite{liu2023visual} in the image domain, and Video LLaMA \cite{damonlpsg2023videollama}, Video Chat \cite{maaz2023video}, and Video ChatGPT \cite{li2023videochat} in the video domain. We use these state of the art Large Multimodal Models to study the prposed new task.

\noindent
\textbf{Prompting techniques}
The emergence of large language models (LLMs) and multimodal models has led to the development of  techniques to enhance their zero-shot abilities. Notable advancements include the Chain of Thought (COT) prompting by Wei et al \cite{NEURIPS2022_9d560961}, Automatic Chain of Thoughts \cite{zhang2022automatic} and the Self-Consistent Chain of Thought \cite{wang2022self} Zhang et al. \cite{zhang2023multimodal} further evolved this concept into the Multimodal Chain of Thought, which incorporates both textual and visual data. Wang et al. \cite{wang2022self} refined the original CoT approach using the self-consistency criteria. Yao et al. \cite{yao2023tree} and Long \cite{long2023large} further proposed through the Tree of Thought. The Graph of Thought by Liu et al. \cite{liu2023graphprompt} expanded on these ideas. Incorporating examples for few-shot learning scenarios has also been shown to improve LLM performance \cite{touvron2023llama, brown2020language} which have been further enhanced upon by \cite{liu2021generated,NEURIPS2020_6b493230,paranjape2023art,zhou2022large}. We analyze these existing techniques capabilities to induce reasoning abilities in LMM's and compare with our proposed method.

\noindent
\textbf{Reasoning abilities of  LLM's} Web et al. \cite{webb2023emergent} showed that models like GPT-3.5 and GPT-4 have considerable analogical reasoning abilities, while Liu et al. \cite{liu2023evaluating} highlighted their limitations with out-of-distribution data and complex tasks. Malkinski et al. \cite{MALKINSKI2023713} analyzed deep models of analytical reasoning on Raven's Progressive Matrices \cite{webb2023emergent}. The Visual Question Answering (VQA) field has seen significant contributions from studies like \cite{Zhang_2023_ICCV}, \cite{Marino_2021_CVPR}, \cite{NEURIPS2018_96ea64f3}, and \cite{Anderson_2018_CVPR}, enhancing VQA solutions. Research by Xue et al. \cite{Xue_2023_CVPR}, Hafner et al. \cite{hafner2019learning}, Finn et al. \cite{finn2017deep}, Chang et al. \cite{chang2016compositional}, Burda et al. \cite{burda2018large}, Babaeizadeh et al. \cite{babaeizadeh2021fitvid}, and Agrawal et al. \cite{agrawal2016learning} has been pivotal in advancing how deep models understand dynamic visuals. To the best of our knowledge LMM's ability to reason over unintentional videos has not been addressed in existing works. 
\noindent
\textbf{Hallucination in LLM's}: Hallucination in foundational models refers to the creation of inconsistent responses. Mckenna et al. \cite{mckenna2023sources} investigated the origins of hallucinations in LLMs, while Yao et al. \cite{yao2023llm} drew comparisons between these hallucinations and adversarial examples. Wang et al. \cite{wang2023evaluation} extended this research to Large Vision Models, examining hallucinations in the visual domain. To address hallucination challenges, Dhuliawala et al. \cite{dhuliawala2023chainofverification} and Manakul et al. \cite{manakul2023selfcheckgpt} introduced self checking and self verification to generate consistent responses. In this work, we use hallucinations to improve the models reasoning capability with the help of multi-step navigation.

\begin{figure*}
    \centering
    \includegraphics[width = \textwidth]{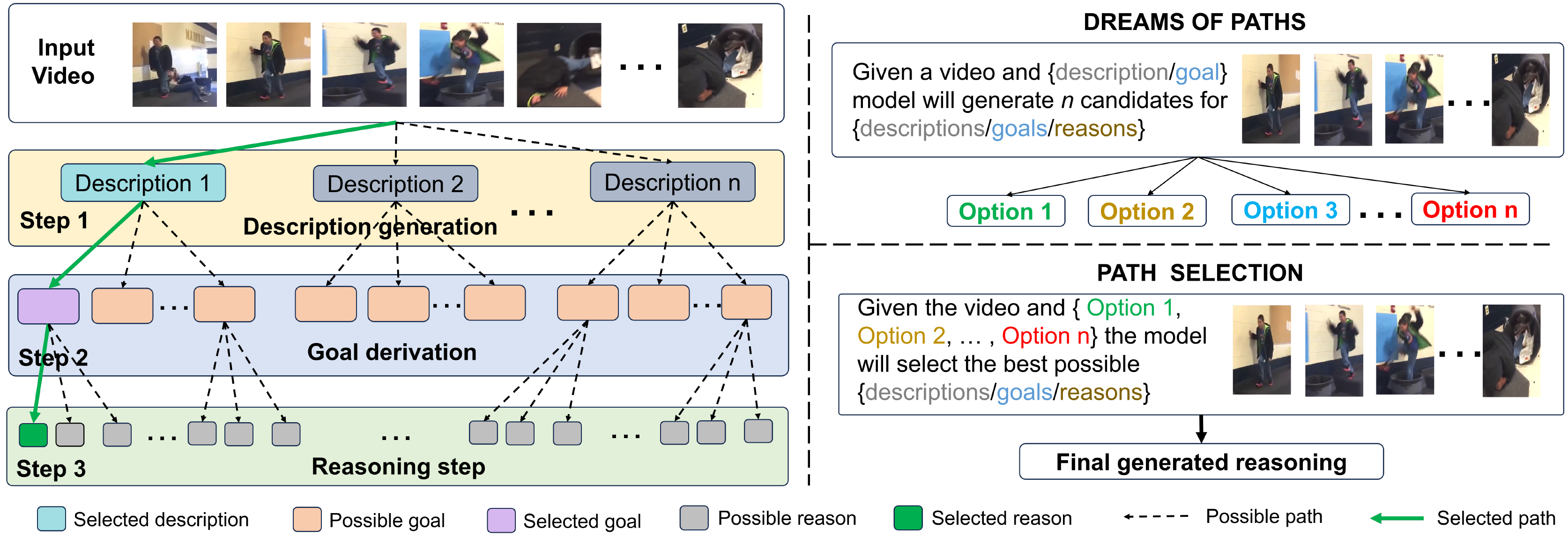}
    \caption{\textit{\textbf{Overview of the proposed Dream of Thoughts framework:}} The left figure shows an overview of the three-step process with all the possible paths generated by the Large Video Language Model using the video and provided prompts. The right figure describes the Dream of Paths mechanism for generating thoughts to cover the most probable options and the Path Selection mechanism for navigating through the best possible options.  
    }
    \label{fig:method}
\end{figure*}

\section{Method}

\textbf{Problem statement}
We focus on understanding the transition from intentional to unintentional activities in videos under zero-shot setting. Given a model $p()$ which takes a prompt $\mathcal{P}$ and a video $\mathbf{V}$ with $n$ frames as input, the objective is to identify the reasoning $\mathbf{R}$ behind the activity's transition from intentional to unintentional in the video.

\subsection{Background and motivation}
The Chain of Thought (COT) prompting \cite{wei2022chain}  method has been shown to enhance the reasoning abilities of LLMs in large-parameter models. Our preliminary experiments indicate that Large Video Language models face specific challenges due to hallucinations as well as lack of ability to infer relationships between events, which seems to be affecting inference and causal understanding. 
While studying these issues, we observe that repeated trials substantially provide accurate responses occasionally, approximately achieving one correct response out of every few attempts with the CoT prompt. Moreover, in \cite{Newell1959ReportOA,newell1972human} the authors show that humans also interpret problem-solving in a combinatorial manner, using some heuristics to decide from various possibilities. The possibilities at each stage are generated by our prior experience in solving problems, which also generates the plan to solve the problem. Motivated by this, we introduce a multi-step prompting strategy which exploits models hallucinations and attempt to navigate through those hallucinated responses to achieve better reasoning. 

\subsection{Proposed approach}
We introduce Dream of Thought (DoT) prompting to improve the models' ability to generate  the correct response by filtering through multiple responses. It is a multi-step process which consists of three steps to obtain cues consisting of essential components to obtain the reason. Specifically, we first obtain a description of the video and using this as the cue, we generate the goal of the intentional activity in the video, which enables us to reason why the intentional activity is failing.  
An overview of the proposed approach is shown in Figure \ref{fig:method}. 
At each step, DoT generates a range of possible answers (Dreams of Paths) to a given question. We then employ a Multiple Choice Question (MCQ)-style prompt for effective selection of the most appropriate response (Path Selection) to the specific video. This strategy capitalizes on the models' generative capability to provide diverse options, with the MCQ prompt acting as a filter to select the most appropriate output.
Similar strategy has been explored in Tree of Thoughts (ToT) \cite{yao2023tree} mechanism but there are some key differences; 1) ToT requires a scoring mechanism to select the best possible option in each step, whereas, we pose this as MCQ for the model itself, and 2) our proposed DoT utilize cues from different steps as a context for next steps, whereas ToT treats each step as a partial path with no such motive.
DoT consists of three main steps, 1) generating description, 2) goal derivation, and 3) reasoning, which make use of Dream of Paths (DoP) and Path Selection. We will first describe Dream of Paths and Path Selection, and then explain the three steps involved in DoT prompting.

\noindent
 \textbf{Dream of Paths}: At each step, we generate $n$ possible options as a solution to the task in corresponding step. The model $p()$ to generate $n$ candidate solutions $x_i \sim p(x_i|V,\dots)$.
 
  \noindent
 \textbf{Path selection}: After obtaining $n$ possible solutions to our problem, we then propose the task as a MCQ form problem where the model has to select one out of n possible solutions: $x \sim p(x|x_1,\dots,x_i,P_s,V)$ using a prompt $P_s$, {``\textit{The list of possible descriptions/goals/reasons for the video are given as (descriptions/goals/reasons). Select the most appropriate descriptions/goals/reasons.}"}
 
 \noindent
 \textbf{Generating description (\textbf{\textit{$\mathcal{D}$}})}: In the first step, we generate $n$ concise summaries of the video content using a prompt: $d_i \sim p(d_i|P_d,V)$,where prompt $P_d$ is {``\textit{Summarize the video action and infer the list of objects exhaustively, from the relevant visual context to the activity occurring in the video.}"}. Following this, we engage in the Path Selection step to derive the most accurate description of the video: $d \sim p(d|d_1,d_2,\dots,d_n,V,P_s )$.

  \noindent
\textbf{Goal derivation (\textbf{\textit{$\mathcal{G}$}})}: Using the summary, we derive $n$ possible intended activity to be executed within the context of this video using a prompt: $g_i \sim p(g_i|d,V,P_g)$, where prompt $P_g$ is given as {\textit{``If the summary of the given video is <video summary>, logically infer the most probable intention of the actions being attempted in this video.}"}. We then perform the Path Selection step to obtain the best possible description for the video: $g \sim p(g|g_1,g_2,g_n,P_s,V,d)$.

  \noindent
 \textbf{Reasoning step (\textbf{\textit{$\mathcal{R}$}})}: Utilizing the information pertaining to the intended activity, we generate a set of $n$ probable factors that could have potentially hindered the successful completion of the aforementioned task: $r_i \sim p(r_i|V,g,P_r)$, using a prompt $P_r$, {``\textit{The goal of the intended activity taking place in the given video is described as: (goal), provide a visual description of the event that leads to the failure to perform the activity with the greatest probability.}"} This step is again followed by the Path Selection step to obtain the best possible description for the video$r \sim p(r|r_1,r_2,r_n,P_r,V,g)$.

 \begin{algorithm}[t!]
 \small
\caption{Dream of Thoughts (DoT)}
\hspace*{\algorithmicindent} \textbf{Input}: Model $\mathcal{M}$, video $V_i$ \\
\hspace*{\algorithmicindent} \textbf{Output}: Reasoning ${R}$ 
\begin{algorithmic}[1]
\State $P = [P_d, P_g, P_r]$ \Comment{Define prompts for reasoning}
\State $c = []$ \Comment{Initialize empty list c for storing context}
\State $n = N$ \Comment{Set $n$ to number of options to be generated}
\State $P_s = Selection Priompt$ \Comment{Set the selection prompt}
\For{$j$ in $P$}
    \State $c_i = []$ \Comment{Initialize empty list $c\_i$}
    \For{$i = 1$ \textbf{to} $n$}
        \State $c_i \mathrel{+}= model(c \mid P_j, V, c)$ \Comment{Update $c\_i$ with model output}
    \EndFor
    \State $c \mathrel{+}= model(c \mid c_i, c, V, P_s)$ \Comment{Update $c$ with model output}
\EndFor
\State $R$ $= c[-1]$ \Comment{Set reason to the last element of $c$}
\end{algorithmic}
\end{algorithm}   

\begin{table*}[t!]
\centering
\small
\begin{tabular}{l|c|c|c|c|c|c|c|c}
\hline
Models        & \multicolumn{4}{c|}{MCQ}                                   & \multicolumn{4}{c}{FIB}                                   \\\hline
              & \multicolumn{2}{c|}{w goal} & \multicolumn{2}{c|}{w/o goal} & \multicolumn{2}{c|}{w goal} & \multicolumn{2}{c}{w/o goal} \\\hline
              & $rm_{MCQ}$        & $rm_{LLM}$        & $rm_{MCQ}$         & $rm_{LLM}$         & $rm_{FIB}$        & $rm_{LLM}$         & $rm_{FIB}$         & $rm_{LLM}$        \\\hline
Video ChatGPT & 0.303        & 0.667       & 0.240         & 0.457        & 0.352        & 0.648       & 0.222         & 0.519        \\
Video LLaMA   & 0.105        & 0.092       & 0.099         & 0.054        & 0.383        & 0.139       & 0.167         & 0.206        \\
Video Chat    & 0.315        & 0.204       & 0.278         & 0.067        & 0.337        & 0.226       & 0.215         & 0.214        \\
Video LLaMAv2 & 0.134        & 0.072       & 0.040         & 0.067        & 0.184        & 0.059       & 0.293         & 0.214       \\
\hline
\end{tabular}
\caption{\textbf{\textit{Reasoning capability of existing models:}} Performance evaluation of existing models on multiple-choice questions (MCQ) and fill-in-the-blank (FIB) style prompting. We analyze both scenarios, prompts with and without goals. MCQ setup consist of four questions, 1 ground truth, 2 random and `\textit{None of the above}'. 
}
\label{tab: Prompt_vartions}

\end{table*}

\subsection{Evaluation and metrics} 
We perform comparison of the responses with the ground truth reasons at both high and low level context. For high level context analysis, we aim to match underlying reasons provided by the model with the ground truth reasoning. For this, we introduce the $rm_{LLM}$ metric. For low level contextual analysis we measure how accurately the model can predict specific attributes of the reason such as subject, verb and object. We propose two metrics for this, $rm_{MCQ}$, and $rm_{FIB}$. Leveraging keyword-based metrics, we can more precisely assess the presence of hallucinations in these models. Specifically, if the keywords are absent, it suggests that hallucination may have occurred, where the keywords have either been replaced by synonyms or include hallucinatory details not originally present.
\noindent
1) \textbf{Low level context evaluation}: The ground truth encompasses subject, object, and verb components extracted from the ground truth, denoted as $s_i$ for the $i^{th}$ video.
Our evaluation revolves around the identification of these ``keywords" within the predicted responses. This evaluation is applied when the reasoning task is framed as either a multiple-choice question (MCQ) task, or a fill-in-the-blanks task. We experimented with existing metrics for generated text evaluation such as BLEU and Sacre BLEU, but these metrics were unable to match the responses providing most of the scores close to 0 therefore we do not use these metrics. 

\noindent
1.1) \textit{MCQ evaluation:}
    For MCQ style task, since we provide the ground truth option as one of the options and rest of the options are unrelated, the presence of keywords in the response provides a reasonable estimate of how correct the answer is and also allows us to judge the accuracy of the output. The $rm_{MCQ}$ accuracy is obtained as,
    \begin{equation}
        rm_{MCQ} = \Sigma_{i=1}^{N} \mathbf{1}[s_i \in pred_i]
    \end{equation}
    where $pred_i$ is the prediction given by the model for the $i^{th}$ video in the dataset. Here N is the total number of samples and $pred_i$ is the prediction provided by the model for the $i_{th}$ video. 

\noindent
1.2) \textit{Fill-in-blank evaluation:}
    In FIB style task since we are removing one of the possible keywords which has to be completed by the model we evaluate the number for keywords model is able to output correctly. We remove $s_i$ from the ground truth reason $gt_i$.
    \begin{equation}
        rm_{FIB} = \Sigma_{i=1}^{N} \Sigma_{x_j \in s_i} \frac{\mathbf{1}[x_j \in pred_i]}{len(s_i)},
    \end{equation}
    Here N is the total number of samples, $pred_i$ is the predicted made by the model for the $i_{th}$ video.

    \noindent
    2) \textit{Reasoning evaluation:}
     Finally, we evaluate the response provided by the models and match it with the ground truth answer. We make us of using GPT-3.5 for matching the generated and ground truth reason. This evaluation allows us to compare whether the output contains the event which occurs in the ground truth reason. We evaluate the same video five times and report the average score of each video as the $rm_{LLM}$ and the standard deviation of scores per question as $std$. 

\section{Experiments}

\label{sec:formatting}
\noindent
\textbf{Datasets} We performed our experiments on two different datasets, OOPs \cite{Epstein_2020_CVPR} and UCF-Crimes \cite{sultani2018real}. \textbf{OOPs:} We conduct detailed experimental analysis using the validation subset of the OOPs dataset. This subset comprises 3,500 YouTube videos, each portraying a variety of failures in diverse real-world scenarios. Along with this, the OOPs dataset also contains natural language descriptions for each video. These descriptions provide insights into the original intentions behind the videos and the circumstances leading to the deviation from planned actions. 
\textbf{UCF-Crimes}
Further, we also conduct experiments on UCF-Crimes dataset. It consists of long
and untrimmed real-world surveillance videos, with 13 realistic anomalies such as fighting, road accident, burglary,
robbery, etc. We use the validation set of this dataset to evaluate our approach, where we select only anomalous videos. These videos have length ranging from 1-3 minutes and there are a total of 65 videos in this evaluation set. We provide natural language descriptions for the crime occurring in the videos from this new test set to evaluate our approach.

\noindent
\textbf{Baselines and models } 
For the evaluation and benchmark, we utilize the officially released versions of several state-of-the-art models, namely Video ChatGPT \cite{maaz2023video}, Video LLaMA \cite{damonlpsg2023videollama}, Video Chat \cite{li2023videochat}, and Video LLaMAv2 \cite{damonlpsg2023videollama}. Along with these video-based models, we also use image based model, Open Flamingo \cite{alayrac2022flamingo}. These models serve as comprehensive baselines in our analysis. Further, we also evaluate different prompting strategies including standard prompting and CoT prompting. Each of these models is built upon the LLaMA-7b billion language model, endowing them with substantial capabilities in text generation from video inputs. 
For the proposed DoT approach, we use Video ChatGPT in all our experiments. 

\begin{figure*}[t!]
    \centering
    \includegraphics[width = \textwidth]{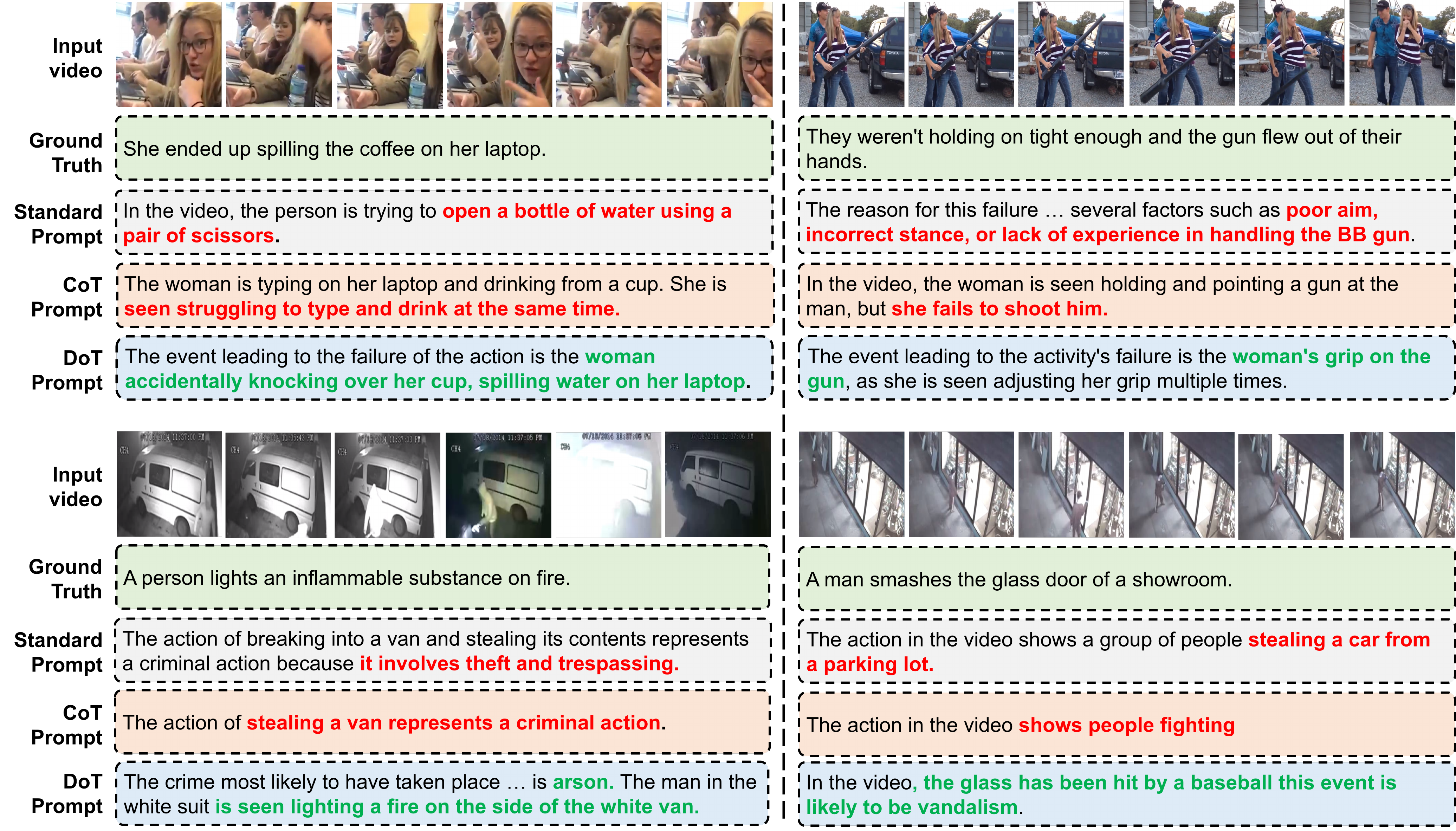}
\caption{\textbf{\textit{Qualitative evaluations:}} We show some samples for qualitative analysis of the proposed DoT prompting compared with CoT  and standard prompting. First row illustrates examples from OOPs dataset and the second row refers to examples sampled from UCF-Crimes dataset.}
    \label{fig qualitative}
\end{figure*}

\subsection{Quantitative results}
We first analyze the reasoning capability of existing LMMs for explaining reasoning behind unintentional activities in videos. Here we explore two different prompting setups, 1) multiple choice questions (MCQs), and 2) fill-in-the-blanks. In MCQ style prompting with n = 3 options (more details in supplementary), we presented several options along with ground truth and prompted the model to select the correct reasoning for the failure. This is evaluated using $rm_{MCQ}$ and $rm_{LLM}$ metrics.
In the second setup, we use the ground truth reasoning and randomly remove subject, object or verbs from the sentence and prompt the model to fill in the missing words. This is evaluated using $rm_{FIB}$ and $rm_{LLM}$ metrics. 
The performance of studied models for MCQ and FIB style prompting is shown in Table \ref{tab: Prompt_vartions}. For both, we experimented with two variations, one where the goal is also provided along with the prompt and the other where goal is not provided. Video ChatGPT shows consistently better performance on both FIB and MCQ prompts for all three metrics with and without goal. Video LLaMA and LLaMAv2 show significantly worse performance on MCQ as compared to FIB-style prompts on $rm_{MCQ},rm_{FIB}$ and $rm_{LLM}$. Video Chat shows similar performance on $rm_{MCQ}$ and $rm_{FIB}$ but $rm_{LLM}$ for FIB is higher in non-goal setting and similar in with goal setting. Based on this analysis, we experimented with mostly Video ChatGPT for proposed DoT prompting technique.

\begin{table}[t!]
\footnotesize
\centering
\begin{tabular}{l|c|c|c|c}
\hline
Dataset & \multicolumn{2}{|c|}{OOPs} & \multicolumn{2}{|c}{UCF-Crimes} \\
\hline
Model    & $rm_{LLM}$ & $std$ & $rm_{LLM}$ & $std$ \\
\hline
\hline
Open Flamingo & 0.154 & 0.128 & 0.035 & 0.047 \\
Video LLaMA & 0.026 & 0.048 & 0.075                         & 0.072                       \\
Video Chat   & 0.064    & 0.156 & 0.082    & 0.143    \\
Video LLaMA2  & 0.053 & 0.089  & 0.081 & 0.089 \\
Video ChatGPT   & 0.242                        & 0.217 & 0.247                        & 0.171                        \\
\hline\hline
CoT      & 0.236                        & 0.182                        & 0.271                        & 0.182                        \\
DoT    & 0.279                       & 0.199                    & 0.291                       & 0.160                        \\
\hline
\end{tabular}
\caption{\textbf{\textit{Performance evaluation:}} A comparison of existing methods with proposed DoT prompting on OOPs and UCF-Crimes dataset. We show both $rm_{LLM}$ and standard deviation ($std$) across five trials. CoT refers to Chain of Thoughts and DoT refers to the proposed prompting strategy using VideoChatGPT model. 
}
\label{tab:ucf_reasoning_results}
\end{table}

\begin{figure}[t!]
    \centering
    \includegraphics[width = \linewidth]{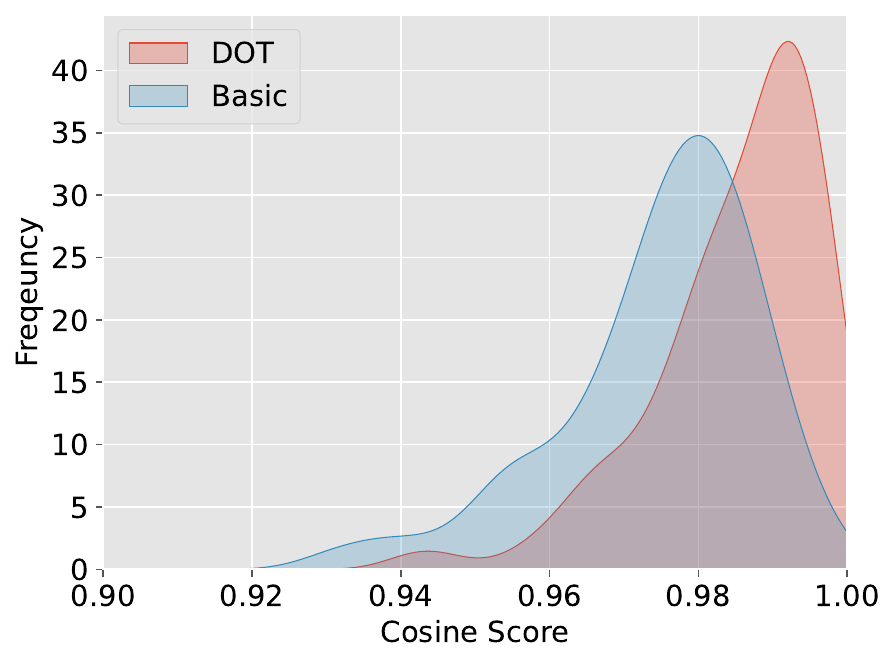}
    \caption{Distribution of cosine similarity between ground-truth and the DoT as well as basic prompt.}
    \label{fig:hall_prompt}
\end{figure}

Next, we evaluate the existing and proposed methods for generating the complete reasoning. We evaluate both CoT and DoT prompting for Video ChatGPT as it was the best performing model in our preliminary experiments. This is evaluated using $rm_{LLM}$ metric along with standard deviation in responses $std$, which attempts to measure degree of hallucinations in the response. 
The evaluation for all the models is shown in Table \ref{tab:ucf_reasoning_results} for both OOPs and UCF-Crimes dataset.
We can observe that the proposed DoT prompting demonstrate benefits over existing methods 
surpassing both the standard and CoT prompts. DoT outperforms Basic prompts by $\sim$ 4$\%$ 
Furthermore,  Video ChatGPT outperforms Video LLaMA, Video LLaMAv2, and Video Chat models when subjected to basic prompts. Similar results can be observed for UCF-Crimes dataset. 
\begin{figure}[t!]
    \centering
    \includegraphics[width = \linewidth]{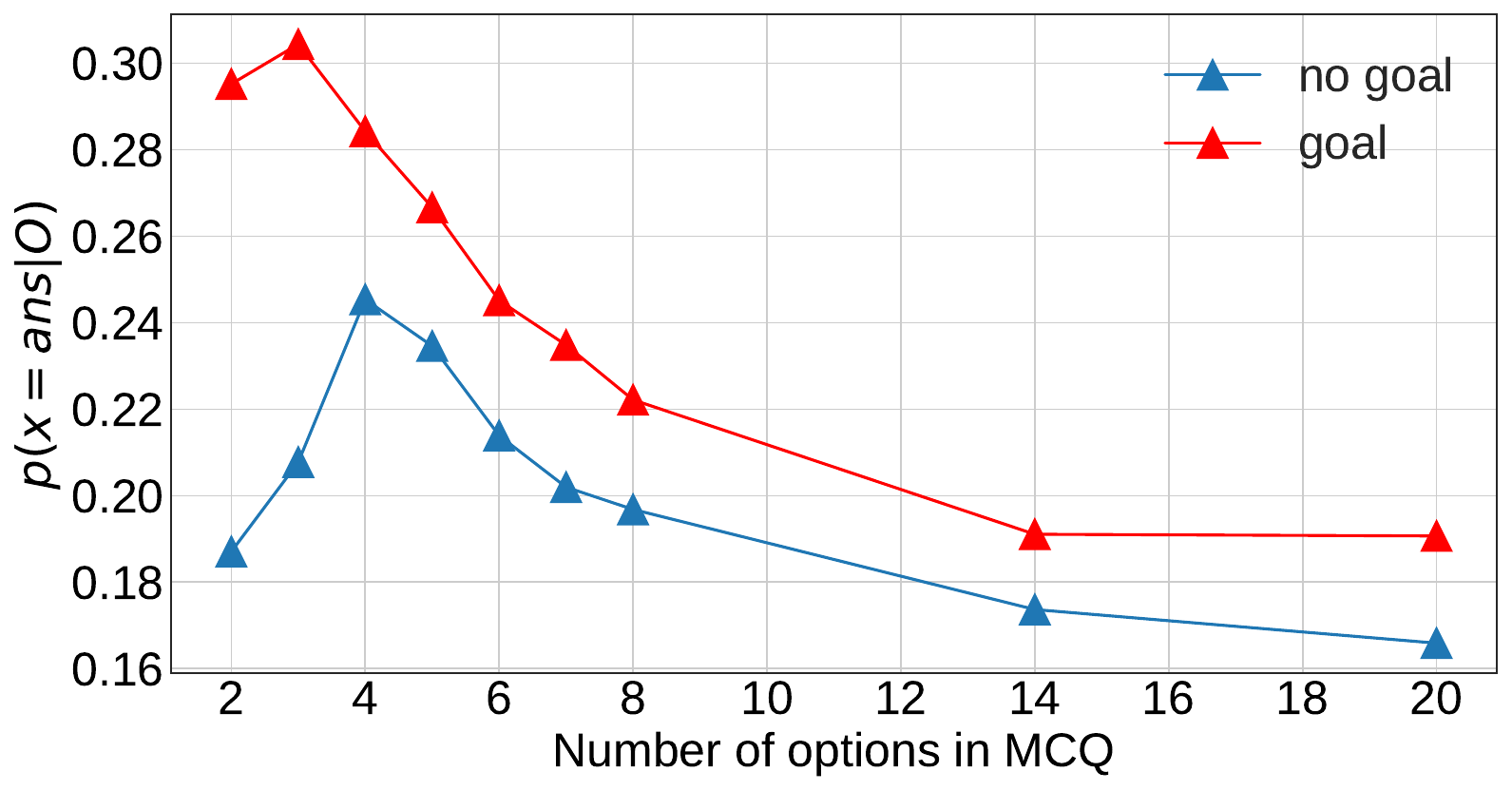}
    \caption{\textbf{\textit{Effect of number of options:}} Variation of $p(x=ans|O)$ on reasoning task proposed as MCQ style query, with varying number of present in a MCQ question, where $p(x=ans|O) = 1 iff rm_{mcq}>=0.8$ else $p(x=ans|O) = 0$. Here $O$ refers to the options presented in the MCQ.
    }
    \label{fig:MCQ}
\end{figure}

\noindent
\textbf{Analyzing hallucinations:}
We provide insights into the standard deviation of scores across individual questions. High standard deviation implies inconsistent answers and substantial model hallucinations. Conversely, a low standard deviation, coupled with low accuracy, suggests consistent but incorrect responses, while a low standard deviation with high accuracy indicates consistent and correct answers. From Table \ref{tab:ucf_reasoning_results} we can observe that DoT has lower $std$ score than basic prompts by $\sim$ 0.02 whereas it is comparable to that of COT, whereas CoT maintains low uncertainty but struggles to consistently achieve high scores when compared to DoT. Additionally, in Figure \ref{fig:hall_prompt} we can see that the outputs obtained from DoT prompt display a consistently higher cosine similarity score to ground truth reason as compared to the output obtained from standard prompts (More details in supplementary).

\subsection{Qualitative Results }
We present qualitative results on the OOPs and UCF-Crimes dataset in Figure \ref{fig qualitative}. We can observe that DoT prompting is generating better reasoning for action failures as well reasoning behind the the activity being anomalous in videos, compared to Standard and CoT prompting. The DoT method is better aligned with ground truth reasoning, showcasing its capability across diverse activities such as typing, shooting an air gun. These activities highlight different success scenarios: ongoing success in working, and instant success in air gun shooting. It also demonstrates its effectiveness to identify a wide range of crimes like arson and vandalism showcasing its generalizability.
\begin{figure}[t!]
    \centering
    \includegraphics[width = \linewidth]{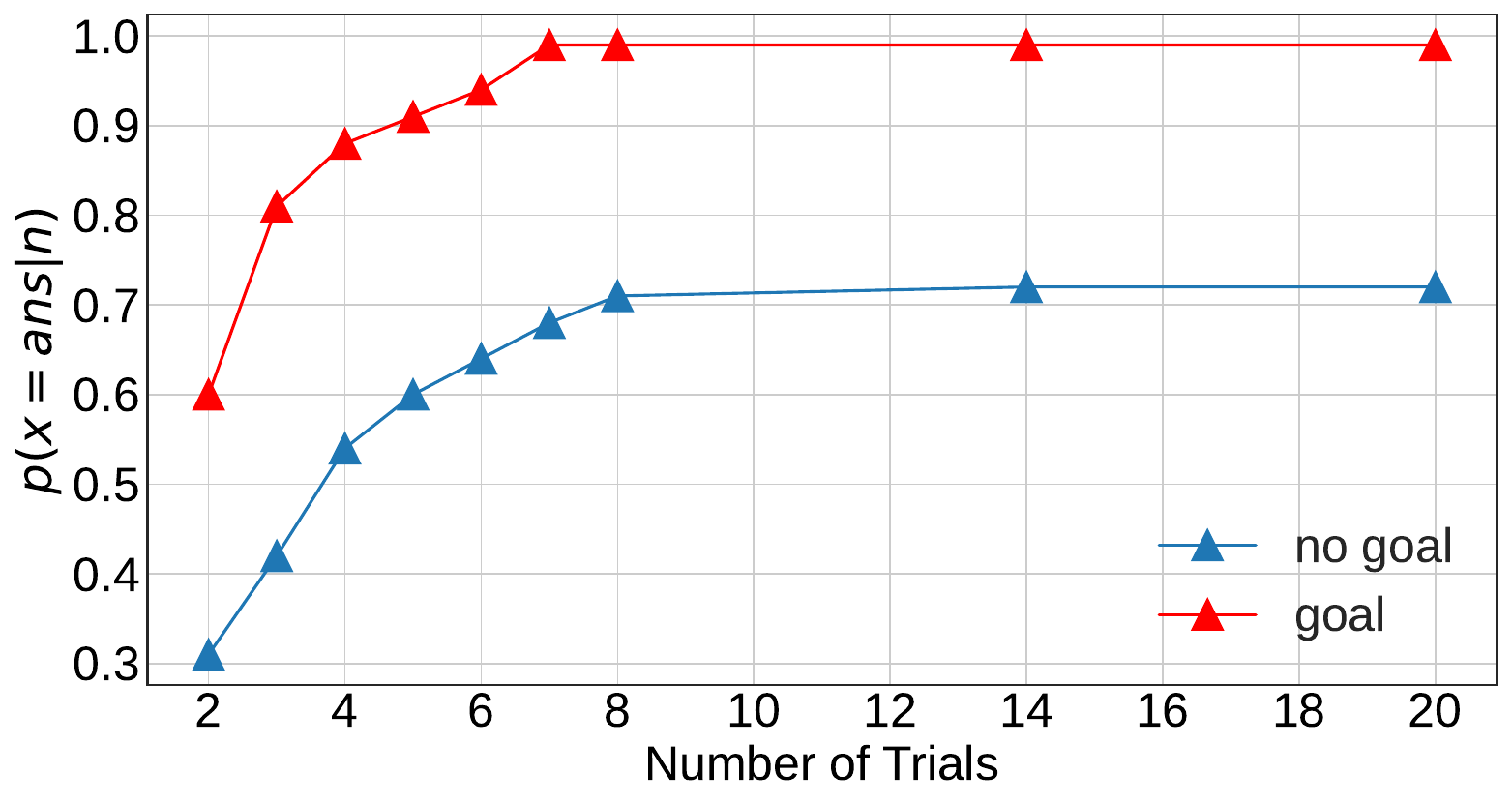}
    \caption{\textbf{\textit{Analyzing number of trials:}} Variation of $p(ans \in x|n)$ on reasoning task proposed as MCQ style query, with $n$ is the number of times prompt has been evaluated using LMM and x is set of n outputs obtained using LMM.
    }
    \label{fig:num_tries}
\end{figure}

\subsection{Ablation studies}
We conduct ablation studies to assess the impact of prompt variations on both accuracy and the presence of hallucinations these ablations studies aid in evaluating the efficacy of each individual step within our proposed DoT prompting methodology. 

\noindent
\textbf{Effect of number of options:}
In MCQ-style question answering, we explore how varying the number of options in MCQs impacts models performance. As shown in Figure \ref{fig:MCQ}, we initially observe some gain of 3$\%$ and 6$\%$ for with and without goal settings respectively which is followed by a noticeable reduction of 12$\%$ in the average $rm_{MCQ}$, when the number of options is increased in both scenarios—with and without a defined goal. 
We hypothesize that the first increment is due the fact that more options allow the model to generate better options with more probability as shown in Figure \ref{fig:num_tries}, but then the performance decreases. This decrease is likely due to the broadening of the model's search space, resulting in more inaccuracies. The score becomes almost constant after 14 options for both with goal and without goal cases.
\\
\begin{table}[]
    \footnotesize
    \centering
    \begin{tabular}{l|c|c|c|c}
    \hline
     & \multicolumn{2}{|c}{with goal} & \multicolumn{2}{|c}{w/o goal}\\
    \hline
    Model   & $rm_{LLM}$ & $std$ & $rm_{LLM}$ & $std$ \\
    \hline
Video ChatGPT                        & 0.621                        & 0.213 & 0.242 & 0.217                        \\
Video LLaMA                                                  & 0.337                         & 0.261 & 0.026 & 0.048                       \\
Video Chat                            & 0.205          & 0.301   & 0.064 & 0.156     \\
Video LLaMA2   & 0.033 & 0.032 & 0.053 & 0.089\\
\hline
    \end{tabular}
    \caption{\textbf{\textit{Effect of goal:}} Performance comparison of models on reasoning with provided goals.
    }
    \label{tab:reason_goal}
\end{table}

\begin{figure}[t!]
\centering
    \includegraphics[width = \linewidth]{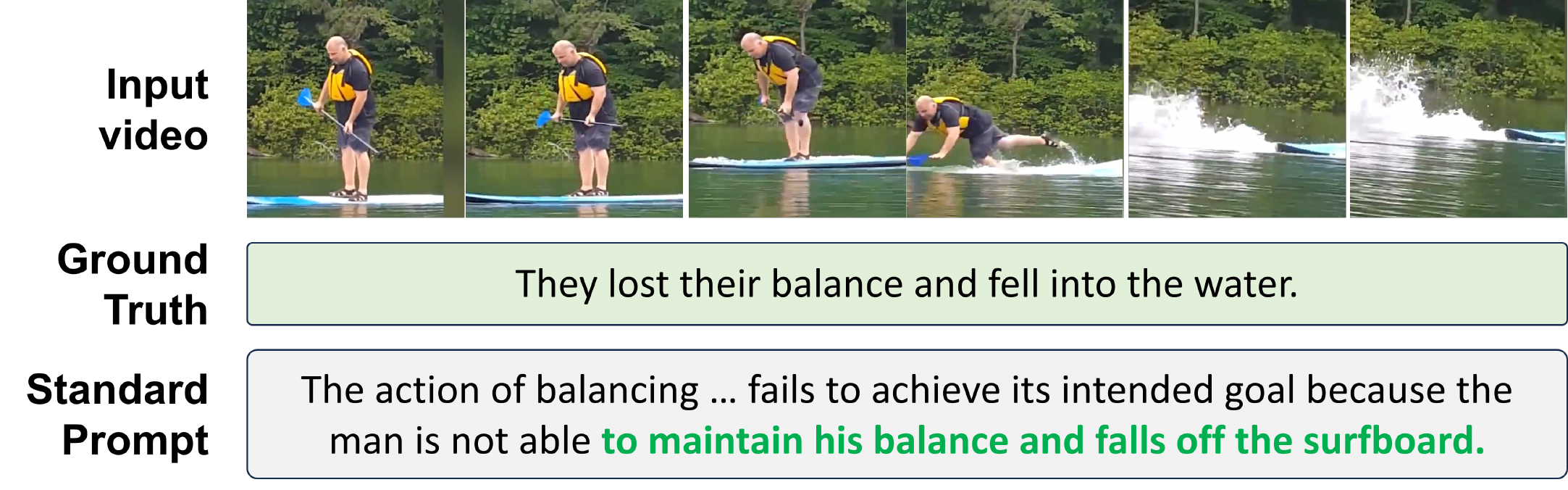}
    \caption{\textbf{\textit{Role of visual information:}} We observe some interesting scenarios where the model using a standard prompt with goal of the video provided is able to infer the correct reasoning without any video frames.
    }
    \label{fig: zero_frame}
\end{figure}

\noindent
\textbf{Effect of goal:} 
Humans demonstrate an impressive ability to comprehend the reasoning behind actions when guided by contextual information. In this experiment, we introduce the goal of the attempted action as a part of the prompt. For this, we construct the prompt as {Prompt: \textit{``If the goal of the activity occurring in the video is (goal). Explain the reason behind the failure to achieve the desired goal.}"}.Analysis of the results, as presented in Table \ref{tab: Prompt_vartions} and Table \ref{tab:reason_goal}, 
reveals that the inclusion of goal enhances the reasoning capabilities of these models. We can see that the presence of goal increases the $rm_{LLM}$ by 0.4 in Video ChatGPT and by 0.2 $\sim$ 0.3 for Video Chat and Video LLaMA models, whereas Video LLaMAv2 seems to perform worse in both conditions. 

\noindent
\textbf{Effect of Dream of Paths:}
We evaluated the effectiveness of Dream of Paths by modifying the prompt to exclude the Dream of Paths step for both descriptions and goals. The results, as shown in Table \ref{tab:DOT_abalations}, reveal that removing this (DoT(w/o des)) leads to a significant decline in performance. This decrease can be attributed to the reliance on inaccurate descriptions for subsequent steps like goal determination and final reasoning, resulting in incorrect overall outcomes. Furthermore, generating a single option for both description and goal (DoT(w/o goal des)) shows marginally better performance compared to DoT(w/o des), yet it falls short of the complete DoT method. 

\noindent
\textbf{Effect of Path Selection}
We compared our Path Selection procedure used in against the DoT($rm_{FIB}$) approach, where we select the option with the highest $rm_{FIB}$ at each stage, ensuring that the option mentioning the most objects involved in the video is chosen. Our results, as detailed in Table \ref{tab:DOT_abalations}, show that using the FIB method, while resulting in a lower $std$, achieves a slightly lower performance compared to the base DoT by 2$\%$.

\begin{table}[]
\centering
\small
\begin{tabular}{l|c|c}
\hline
Model                            & $rm_{LLM}$ & $std$ \\
\hline \hline
CoT                              & 0.237     & 0.182 \\
\hline
DoT(w/o des)  &    0.180       &   0.153       \\
DoT(w/o goal,des) & 0.221          &  0.182        \\
DoT($rm_{FIB}$) & 0.260                       & 0.183                        \\
\hline
DoT                        & 0.279     & 0.199    \\
\hline
\end{tabular}
\caption{\textbf{\textit{Ablation Analysis of the DoT Prompt}}.DoT(GPT):final path selection is performed using GPT-3.5. DOT(w/o des) refers to the case when we directly obtain description. Similarly, in DoT(w/o goal, des) we directly obtain goal and description. In DoT($rm_{FIB}$) the path selection is performed using $rm_{FIB}$. 
}
\label{tab:DOT_abalations}

\end{table}

\begin{figure}[t!]
    \centering
    \includegraphics[width = \linewidth]{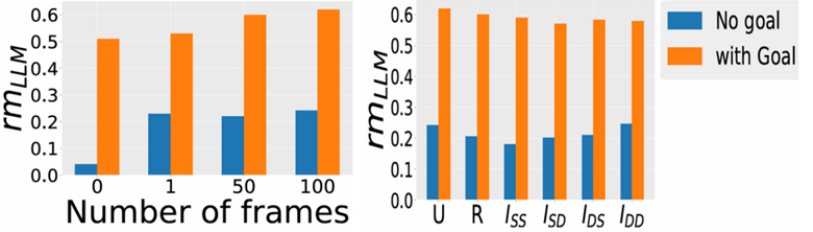}
    
    \caption{\textbf{\textit{Effect of number of frames and sampling strategy:}} The left plot shows the effect of varying the number of sampled frames on $rm_{LLM}$ for reasoning task. In the right plot we show effect of various frame sampling techniques in videos: U(uniform sampling), R(random sampling), $I_{SS}$ (sparse sampling from both intentional and unintentional parts), $I_{SD}$ (sparse from intentional, dense from unintentional), $I_{DS}$ (dense from intentional, sparse from unintentional), and $I_{DD}$ (dense sampling from both intentional and unintentional parts)
    }
    \label{fig:num_frames}
\end{figure}

\subsection{Analysis}

\noindent
\textbf{Number of video frames:} We conduct an analysis on the effect of number of video frames to investigate their impact on models performance. We vary the number of frames, ranging from 0 to 1, 50, and 100 frames. Our observations, as depicted in Figure \ref{fig:num_frames}, reveal that the model's performance remains relatively stable concerning the number of frames but experiences a substantial drop when no frames are provided as input. Interestingly, for some scenarios (Figure \ref{fig: zero_frame}) when merely a goal is provided to the model, it manages to achieve a significantly high $rm_{LLM}$ using only the goal as information about the video, which shows that it utilizes textual conditioning more efficiently than visual modality. 

\noindent
\textbf{Sampling strategy:}
Additionally, we explore variations in the frame sampling strategy, ranging from uniform and random sampling to importance sampling. Importance sampling involves selectively sampling frames sparsely or densely from the intentional and unintentional segments of the video. To execute importance sampling, we utilize timestamps provided for intentional and unintentional parts of the video with the OOPs dataset, sampling varying numbers of frames from the start of the video to the beginning of the transition, and from the start of the transition to its end. Our findings, presented in Figure \ref{fig:num_frames}, show that sampling strategies do not significantly affect the reasoning capabilities of Video ChatGPT, uniform sampling offers the best overall performance, followed by sampling frames densely from intentional and unintentional parts.  

\section{Conclusion}
In this work, we present a novel task regarding understanding of unintentional activities in videos where we formalize it as a zero shot reasoning task. We first analyze the reasoning capabilities of existing LMM models and prompting techniques and then also propose a novel DoT prompting technique which navigates through hallucinations introduced by LLM's to obtain the reasoning. We propose different metrics to quantify the models performance and also analyze hallucinations of the responses. We further demonstrate that the proposed method outperforms existing prompting techniques.

\section{Guidelines}
\subsection{Limitations} In this work we only explore reasoning where the event that causes the action to fail occurs immediately before the actual failure of the action. We do not consider actions which may cause failure of the action at a later moment in time. 

\subsection{Risks} This research may pose some risk for privacy by being employed extensively for surveillance.

\subsection{Licenses}
OOPs dataset -  Creative Commons Attribution-NonCommercial-ShareAlike 4.0 International License.
Video ChatGPT- Creative Commons Attribution-NonCommercial-ShareAlike 4.0 International License.
LLaMA- LLAMA community license agreement
UCF-Crimes - Creative Commons Attribution-NonCommercial-ShareAlike 4.0 International License.

\subsection{Computation}
All experiments we performed using a single V-100 32 GB GPU with each experiment taking around 10 hours.
{
    \small
    \bibliographystyle{ieeenat_fullname}
    \bibliography{egbib}
}
\newpage
\appendix

\section{Appendix}
\subsection{Cosine similarity}
To obtain the cosine similarity score for Figure \ref{fig:hall_prompt} we prompt the model as the \textbf{Prompt: ``\textit{Given the video goal of the activity occurring in the video as <goal> and reason behind its failure as <reason>}``} and take the embedding obtained from the encoder of Video-ChatGPT model. For ground truth encoding we replace <reason> with the ground truth reason similarly for DoT and Basic prompt with reasoning obtained from using repsective prompts.
\subsection{LLM Evaluation}
We use GPT-3.5 for evaluation using LLM. To obtain the score we prompt GPT-3.5 as 
\textbf{Prompt:} 
\textit{"You are provided with a question,the correct answer and the predicted answer. The question contains information about the task being attempted to be achieved in the video, along with the context about the objects involved in achieving that goal. The correct answer consists of the reasons behind the failure of achieving that objective and information about the objects present during the failure. Your task is to evaluate the correctness of the predicted answer. Here's how you can accomplish the task://
                                {"------"
                                "INSTRUCTIONS: //"
                                "- Focus on the meaningful match of events between the predicted answer and the correct answer.\\"
                                "- Consider synonyms or paraphrases as valid matches.\\"
                                "- Evaluate the correctness and alignment of the predicted answer compared to the correct answer.\\"
                        },\\
                        {\\
                            "role": "user",\\
                            "content":\\
                                "Please evaluate the following video-based question-answer pair:\\"
                                f"Question: {question}\\"
                                f"Correct Answer: {answer}\\"
                                f"Predicted Answer: {pred}\\"
                                "Provide your evaluation only as a yes/no and score where the score is an integer value between 0 and 1, with 1 indicating the highest meaningful match. "
                                "Please generate the response in the form of a Python dictionary string with keys 'pred' and 'score', where value of 'pred' is  a string of 'yes' or 'no' and value of 'score' is in NUMBER, not STRING."\\
                                "DO NOT PROVIDE ANY OTHER OUTPUT TEXT OR EXPLANATION. Only provide the Python dictionary string. "
                                "For example, your response should look like this: {'pred': 'yes', 'score': 0.8}."}}
Where the correct reason is the ground truth reason the question is given as \textbf{\textit{If the <goal> of the action occurring in the given video infer the reason why the action fails to achieve the intended outcome }} and predicted answer is the answer obtained using the respective prompting technique.

\subsection{MCQ Style Prompt}:
To formulate the MCQ style prompt mentioned in \ref{tab: Prompt_vartions} containing n options we first randomly select ground truth reasons behind the failure of actions to obtain n-2 options. In addition to these N-2 options we also provide the ground truth reason for that particular video and None of these option as well. The prompt provided to the model is given as \textbf{\textit{The action occurring in the given video fails.You will be given ${num\_options}$  describing the reasoning behind the failure. The options for this video are given as ${options\_list}$.}} where $num\_options$ is the number of options provided in the MCQ style prompt and  $options\_list$ refers to the list of options provided to the MCQ style prompt. 

\subsection{FIB style prompt}
To formulate the FIB style prompt used in \ref{tab: Prompt_vartions} we first use the ground truth reason behind the failure contain a list of $s$ subjects $v$ verbs and $o$ objects. First we randomly remove $s$ , $v$ and $o's$ and replace it with $\_\_\_$. The sentence obtained after it is \textbf{\textit{They $\_\_\_\_\_\_$ the $\_\_\_\_\_\_$ too high and $\_\_\_$ a $\_\_\_\_\_\_$ $\_\_\_\_\_\_\_$ off.}} Finally we prompt the model with \textbf{\textit{Given the following video complete the following sentence such that the sentence describes the reasoning behind failure of the intended action in the video. The sentence to be completed is <sentence>. Note: Your task is to complete the given sentence where the blanks are indicated by $\_\_\_\_\_$.}}

\subsection{UCF-Crimes Dataset Annotation}
\begin{figure*}
    \centering
    \includegraphics[width = \textwidth]{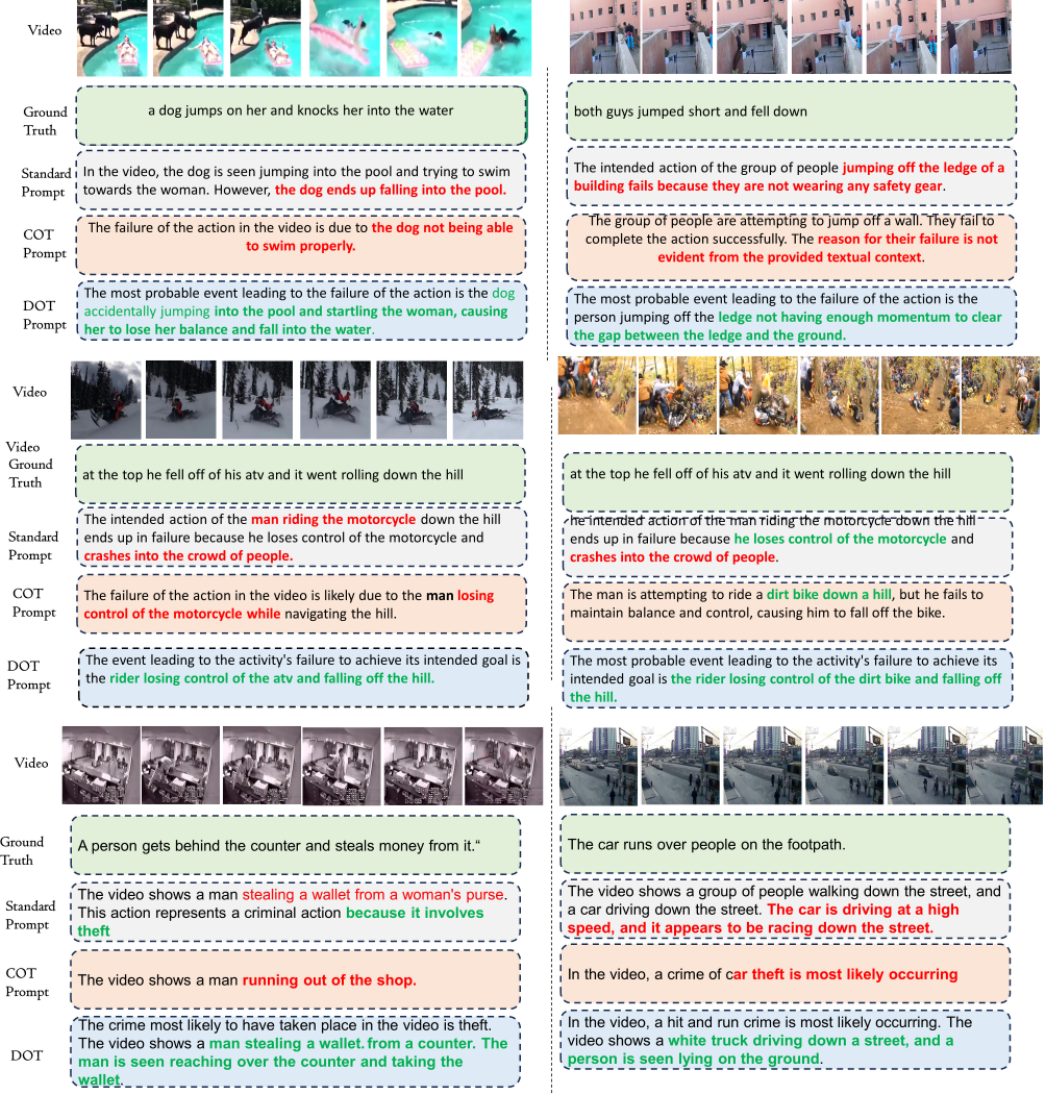}

\caption{We show some samples for the qualitative results of the proposed DOT prompting compared with COT and standard prompting for UCF-Crimes and OOPs dataset.
}
    \label{fig qualitative_Crime_b}
\end{figure*}

UCF-Crimes Dataset does not provide natural language descriptions for the reasoning behind the event occurring the video being a crime. We manually annotate each anomalous video in the validation set by providing information about the actor, who commits the crime , the crime committed in the video and the victim of the crime, if applicable in the video for example in Figure \ref{fig qualitative_Crime_b} in the last row represent examples from UCF-Crimes dataset. From the ground truth annotations we can note the presence of the actor the crime and victim(if present) in each annotation.

\begin{figure*}
    \centering
    \includegraphics[width = \textwidth]{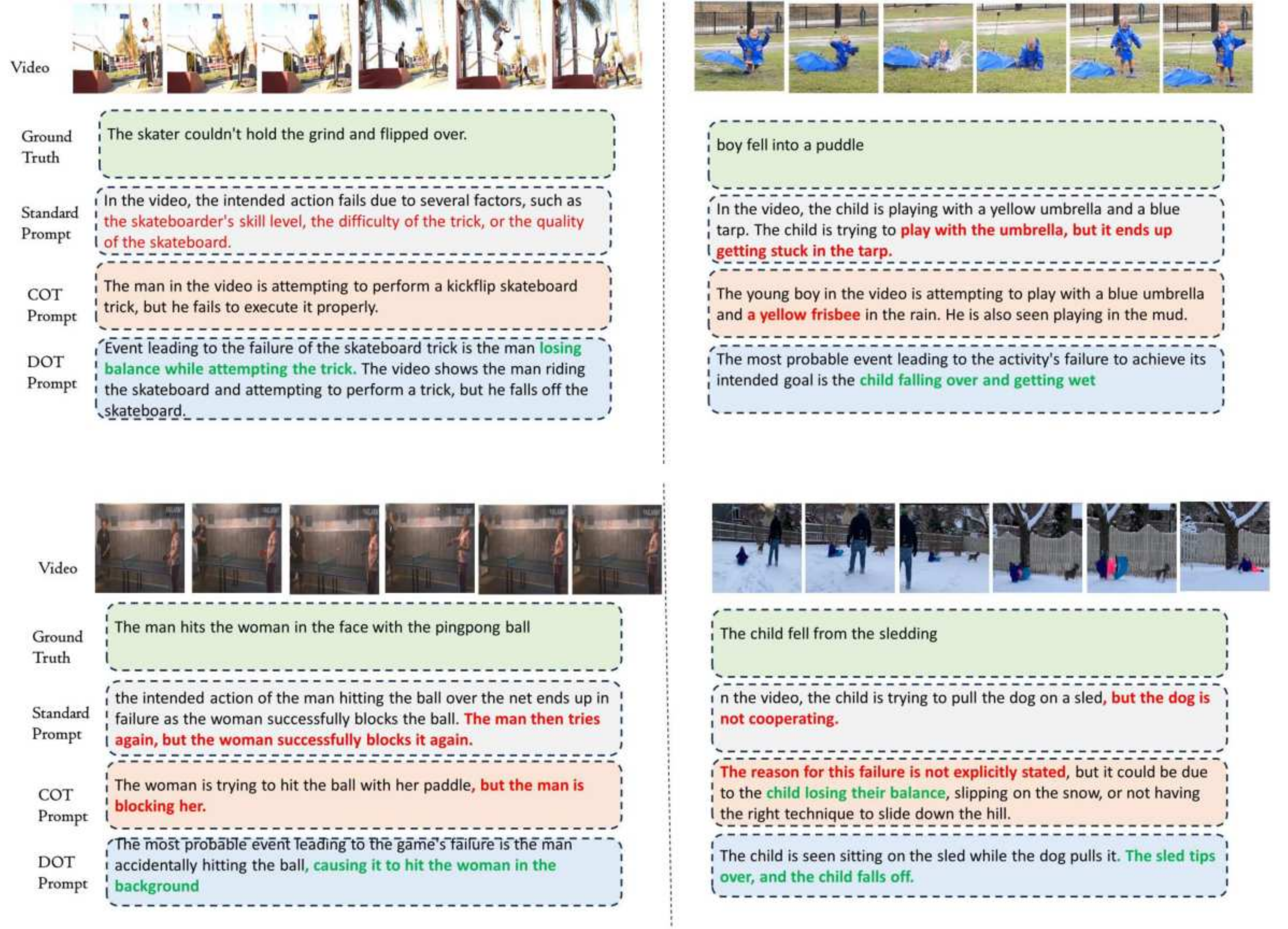}

\caption{We show some samples for the qualitative results of the proposed DOT prompting compared with COT and standard prompting for OOPs dataset.
}
    \label{fig qualitative_Crime_r}
\end{figure*}

\end{document}